\documentclass[conference]{IEEEtran}
\IEEEoverridecommandlockouts
\usepackage{float}
\usepackage{comment}
\usepackage{subcaption}
\usepackage{placeins}
\usepackage{url}

\usepackage{cite}
\usepackage{amsmath,amssymb,amsfonts}
\usepackage{algorithmic}
\usepackage{graphicx}
\usepackage{textcomp}
\usepackage{xcolor}
\def\BibTeX{{\rm B\kern-.05em{\sc i\kern-.025em b}\kern-.08em
    T\kern-.1667em\lower.7ex\hbox{E}\kern-.125emX}}

\newcommand{\TrainPercent}{70\% }
\newcommand{\TestPercent}{30\% }
\newcommand{\CovRegulizer}{1e-4 }

\begin{document}

\title{Hybrid Dynamics Modeling for a Flexible 2-DoF Robotic Arm
}


\author{
\IEEEauthorblockN{
Maciek Popik,
Daniel Yang,
Mahdis Bisheban
}
\thanks{
Maciek Popik, Daniel Yang, Mahdis Bisheban are with the Dept. of Mechanical and Manufacturing Eng at the Schulich School of Engineering, University of Calgary, Alberta, Canada. Emails: \{maciek.popik, daniel.yang2, mahdis.bisheban\}@ucalgary.ca. }
\thanks{
This work was supported by the Natural Sciences and Engineering Research Council of Canada (NSERC), the Government of Alberta, Alberta Innovates, and the Schulich School of Engineering at the University of Calgary. Funding was awarded to Dr. Mahdis Bisheban, Director of the Intelligent Dynamics and Control Lab and Assistant Professor at the University of Calgary.}
}

\maketitle

\begin{abstract}


This paper examines three approaches for modeling the dynamics of a flexible-link 2-DoF robotic arm to address unmodeled dynamics not captured by rigid-body models.
Two physics informed models combine rigid-body dynamics (RBD) formulations with a Gaussian Mixture Model (GMM) to capture residual model errors and linkage flexibility. A kinematics-based regression model serves as a purely data-driven baseline.
Using an open-source dataset, torque predictions are first estimated using Ridge regression on kinematic features, while the physics-based baseline is constructed from published specifications, and ordinary least-squares regression is subsequently used to estimate the same parameter set directly from data. 
Results show that the physics-based parameters yield the poorest accuracy, while regularized and least-squares estimators align more closely with measured torques. Residual analysis and error metrics highlight the limitations of purely parametric models for flexible-link systems and underscore the value of regularization and data-driven identification, supporting developments of semi-parametric residual learning methods. 

\end{abstract}

\section{Introduction and Problem Statement}
Modeling flexible robotic arms is challenging due to the presence of disturbances and unmodeled effects that cannot be realistically captured by rigid-body dynamics (RBD) alone. Model-based methods, which attempt to build a model that fully encapsulates all underlying effects, and model-free methods, which base model correction on system measurements, have been previously explored \cite{FlexibleLinkReviewPaper}. Using these methods, additional sensor inputs or deformation mechanics information is often required to provide a more accurate estimate of torque.

The flexible-link robotic arm  developed at TU Dortmund University (TUDOR) \cite{TUDOR-MERIt-2014}, compensates for link flexibility through direct, strain-based strategies.
Strain gauges embedded on the links provide measurements of elastic deformation, which are decomposed into static and dynamic components. Dynamic strain feedback is used to actively damp flexible oscillations, while static strain is mapped linearly to load-side joint torque, enabling strain-based estimation of flexibility-induced torques \cite{TUDORFractionalStrain} \cite{TUDORStrainFeedback} \cite{TUDORDynamicsIdentify}. Furthermore, demonstrations have shown the implementation of Fiber Bragg Grating (FBG) sensors on industrial robots used to obtain strain measurements, allowing for the replication of methods from experiments \cite{TUDOR-FBG}. Although these methods have proven to be effective, additional strain measurement sensors are required as input for compensation. For many existing robots, this requires the retrofitting of FBG or other strain measurement sensors. In addition to the calibration and temperature sensitivity challenges of FBG sensors, measurement of link strain inherently does not include joint flexibility (including that of the actuating system, gear tooth compliance, and belt stretch) nor out of plane deformations. 


In an effort to eliminate sensor dependency, alternative approaches include fully defining linkage flexure and incorporating it within a torque model through the evaluation of the robot system using finite element methods (FEM) \cite{TUDOR-FEM}. In \cite{TUDOR-FEM}, a finite-element-based flexible link model is proposed in which, unlike strain-based measurement approaches, link flexibility is captured from first principles using structural mechanics. FEM is employed offline to derive a reduced-order flexible dynamics model that can be used efficiently for control and torque prediction without reliance on strain measurements. Although this approach eliminates sensor dependency, the required offline effort is significant, as it necessitates a full FEM analysis with accurate geometric and material properties. Furthermore, actuator dynamics and hardware-related nonlinearities are not explicitly modeled.

Recent inverse-dynamics research has increasingly shifted toward neural-network and physics-informed neural-network (PINN) models, since these methods can achieve strong predictive performance while embedding physical structure to improve learning and control behavior. Nevertheless, purely neural approaches are still commonly criticized for limited interpretability, as the literature characterizes them as black-box models with ``rather uninterpretable internals'' and notes that standard deep networks are black-box representations that do not transparently expose the underlying physics \cite{Pikulinski2024,LutterPeters2023}.

In comparison, simpler rigid-body dynamics formulations augmented with learning-based residual models offer a middle ground between fully physics-based flexible modeling and fully data-driven approaches, which can be used to improve overall model accuracy \cite{ML1} \cite{ML2} \cite{ML3} \cite{ML4}. At the cost of initial training, accurate models can be determined without additional sensors or detailed geometric and mechanical properties of the system. Although Gaussian Process models provide a flexible non-parametric approach for learning residual robot dynamics, their $\mathcal{O}(N^3)$ training cost and $\mathcal{O}(N^2)$ memory requirements can become prohibitively computationally-expensive in robotics applications \cite{ML3} \cite{Chalupka2013GPApprox}. The method proposed in \cite{8978477} adopts a semi-parametric approach, where the nominal dynamics are first expressed using a RBD model, after which the remaining residuals are captured using an Incremental Gaussian Mixture Model (IGMM) with reduced time complexity \cite{IGMM}. The RBD formulation is linear in the inertial parameters, allowing the dynamics to be represented as a regressor matrix and parameter vector, which can be estimated using least-squares techniques. Starting with an initial subset of pre-collected open source data \cite{TUDOR-MERIt-2014}, the semi-parametric model is then updated online as new data arrives, with a consistency transformation enabling parallel training of the GMM \cite{8978477}.


In this paper, three complementary modeling approaches are implemented and evaluated on the MERIt/TUDOR Dataset: TUD01, without payload, available online~\cite{TUDOR-MERIt-2014}, to assess the effectiveness of data-driven, physics-based, and semi-parametric torque prediction strategies. The data-driven approach employs ridge regression to approximate joint torques from kinematic features, the physics-based approach relies on RBD with parameters derived from published robot properties, and the semi-parametric approach identifies RBD parameters using unconstrained linear least-squares regression~\cite{leastsquares}. In all cases, Gaussian Mixture Regression (GMR) is used to model residual dynamics arising from typically unmodeled effects, such as link flexibility. To improve numerical stability, diagonal covariance regularization is incorporated in the residual model~\cite{diagreg}. The study further adapts an online residual learning workflow to the dataset by sequentially updating the parametric component as new data becomes available while applying the consistency transform proposed by~\cite{8978477} to maintain coherence between the evolving parametric model and the learned Gaussian Mixture Model (GMM).
We have released the code for each of the three approaches through the IDCL project Github repository\footnote{Source code available at \url{https://github.com/IDCL-UCalgary/Hybrid-Dynamics-Modeling-of-Flexible-Robotic-Arms}}

\section{Problem Formulation}
To highlight the challenges introduced by link flexibility in robotic arms \cite{FlexibleLinkReviewPaper}, we selected data from a robotic arm system with pronounced flexible-link characteristics, using the publicly available dataset from \cite{TUDOR-MERIt-2014}.
The selected subset of experiments consists of a 3-Degree-of-Freedom (3-DoF) arm where the first actuator remains stationary, reducing the system to a planar 2-Degree-of-Freedom (2-DoF) arm with two active joints.

\begin{figure}[ht]
\begin{center}
\includegraphics[width=0.6\linewidth]{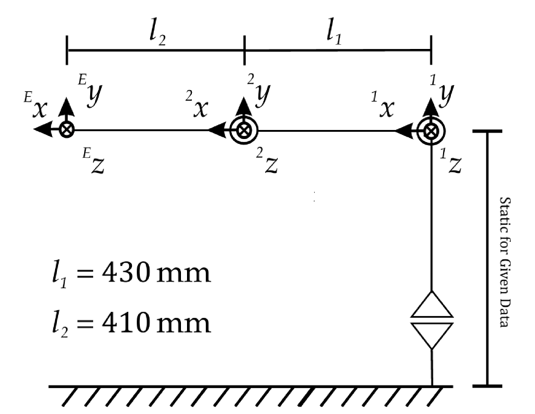}
\caption{Two-degree-of-freedom flexible-link robotic arm showing joint angles and link parameters. Original schematic from \cite{TUDOR-MERIt-2014}, modified to reflect dataset configuration.}
\label{fig:TUDOR_schematic}
\end{center}
\end{figure}

The traditional equation for a 2-DoF robotic arm, assuming rigid body linkages is presented as
\begin{equation} \label{eq:General_Torque_Equation}
    \tau = M(q)\,\ddot{q} + C(q,\dot{q})\,\dot{q} + G(q) + F(\dot{q}),
\end{equation}
where $q\in \mathbb{R}^n$ are joint angles, $\tau \in \mathbb{R}^n$ is the joint torque vector,
$M(q) \in \mathbb{R}^{n \times n}$ is the inertia matrix,
$C(q,\dot{q})$ contains Coriolis and centrifugal terms,
$G(q)$ is the gravity torque vector, and
$F(\dot{q})$ represents joint friction, with $n=2$ for 2 DoF. Expanding on Equation \ref{eq:General_Torque_Equation}, a schematic based on the experimental robotic arm 
has been created in Figure \ref{fig:TUDOR_schematic} to illustrate the RBD parameters considered.

The following physical parameters are used:
$m_1, m_2$ denote the masses of links 1 and 2;
$l_1$ is the length of link 1;
$l_{c,1}, l_{c,2}$ are the centers of mass (COM) of links 1 and 2;
and $I_{c,1}, I_{c,2}$ are the corresponding COM moments of inertia.

For compactness, the shorthand variables $c_1 = \cos(q_1)$
$c_2 = \cos(q_2)$, $s_2 = \sin(q_2)$ and $c_{1,2} =\cos(q_1 + q_2)$ are introduced.
The grouped inertial parameters are defined as
$\boldsymbol{\alpha} = (\alpha_1, \alpha_2, \alpha_3)$
and the gravity parameters as
$\boldsymbol{\beta} = (\beta_1, \beta_2)$
The grouped parameters $\boldsymbol{\alpha}$ and $\boldsymbol{\beta}$ are defined as:
$\alpha_1 = I_{c,1} + I_{c,2} + m_1 l_{c,1}^2 + m_2(l_1^2 + l_{c,2}^2)$,
$\alpha_2 = m_2 l_1 l_{c,2}$,
$\alpha_3 = I_{c,2} + m_2 l_{c,2}^2$,
$\beta_1 = (m_1 l_{c,1} + m_2 l_1) g$,
and $\beta_2 = m_2 l_{c,2} g$,
with $g$ denoting the gravitational acceleration.
Then, the expanded form becomes 

\begin{align} 
M(q) =&
\begin{bmatrix}
\alpha_1 + 2 \alpha_2 c_2 & \alpha_3 + \alpha_2 c_2 \\
\alpha_3 + \alpha_2 c_2   & \alpha_3
\end{bmatrix},\label{eq:Inertia_Matrix}\\
C(q,\dot{q}) =&
\begin{bmatrix}
-2 \alpha_2 s_2 \dot{q}_2 & -\alpha_2 s_2 \dot{q}_2 \\
\alpha_2 s_2 \dot{q}_1    & 0
\end{bmatrix},\label{eq:Coriolis_Centrifugal_Matrix}\\
G(q) =&
\begin{bmatrix}
(\beta_1 + \beta_2) c_1 + \beta_2  c_{1,2} \\
\beta_2  c_{1,2}
\end{bmatrix},\label{eq:Gravity_Vector}\\
F(\dot{q}) =&
\begin{bmatrix}
f_{v1}\dot{q}_1 + f_{c1}\,\mathrm{sgn}(\dot{q}_1) \\
f_{v2}\dot{q}_2 + f_{c2}\,\mathrm{sgn}(\dot{q}_2)
\end{bmatrix}, \label{eq:Friction_Vector}
\end{align}
for the Inertia matrix, Coriolis and centrifugal matrix,  Gravity vector, and Friction vector respectively, where
$f_{v1}, f_{v2}$ are viscous friction coefficients at joints 1 and 2, 
$f_{c1}, f_{c2}$ are Coulomb friction coefficients at joints 1 and 2, and 
$\mathrm{sgn}(\cdot)$ is the signum function.

Substituting the matrices into the RBD equation yields the expanded form:
\begin{align}
\tau
&=
\begin{bmatrix}
\alpha_1 + 2 \alpha_2 c_2 & \alpha_3 + \alpha_2 c_2 \\
\alpha_3 + \alpha_2 c_2   & \alpha_3
\end{bmatrix}
\begin{bmatrix}
\ddot{q}_1 \\ \ddot{q}_2
\end{bmatrix}
\notag\\
&\quad +
\begin{bmatrix}
-2 \alpha_2 s_2 \dot{q}_2 & -\alpha_2 s_2 \dot{q}_2 \\
\alpha_2 s_2 \dot{q}_1    & 0
\end{bmatrix}
\begin{bmatrix}
\dot{q}_1 \\ \dot{q}_2
\end{bmatrix}
\notag\\
&\quad +
\begin{bmatrix}
(\beta_1 + \beta_2)  c_1 + \beta_2  c_{1,2} \\
\beta_2  c_{1,2}
\end{bmatrix}
+
\begin{bmatrix}
f_{v1}\dot{q}_1 + f_{c1}\,\mathrm{sgn}(\dot{q}_1) \\
f_{v2}\dot{q}_2 + f_{c2}\,\mathrm{sgn}(\dot{q}_2)
\end{bmatrix}.
\label{eq:expanded_tau}
\end{align}

Although Equation \ref{eq:expanded_tau} accounts for various frictional losses, flexibility in both joints and linkages is unaccounted for. As previously described, without the use of additional sensors or a deep knowledge and understanding for the robot system's mechanics, these disturbances are difficult to incorporate into the dynamic model.

\section{Methodology}
Within the context of this paper, three models are applied to the open source data \cite{TUDOR-MERIt-2014}. Due to the complexity of explicitly modeling link bending effects within rigid-body kinematics and dynamics, any bending related effects are grouped into the non-parametric torque error component of rigid robotic arm physics \cite{8978477}. This formulation enables a consistent comparison between purely parametric, purely physics-based, and semi-parametric learning approaches. Across all of the methods presented, only the provided joint position, velocity, acceleration and torque measurements are used. Data related to strain measurements is not accessed.

 \begin{table}[H]
\centering
\caption{True Joint Parameters
\cite{TUDOR-Online-Datasheet}}
\label{tab:true_joint_params}
\resizebox{\columnwidth}{!}{%
\begin{tabular}{lccc}
\hline
\textbf{Parameter} & \textbf{Joint 1} & \textbf{Joint 2} & \textbf{Units} \\
\hline
Torque Constant & 4.48\text{E}-02 & 1.29\text{E}-02 & Nm/A \\
Gear Ratio (1:1) & 230 & 246 & -- \\
Rotor and Gear Inertia & 9.41\text{E}-06 & 8.20\text{E}-07 & kgm$^2$ \\
Positive Viscous Friction & 1.86\text{E}-05 & 2.60\text{E}-06 & Nms/rad \\
Negative Viscous Friction & 1.86\text{E}-05 & 2.30\text{E}-06 & Nms/rad \\
Positive Coulomb Friction & 3.34\text{E}-03 & 1.49\text{E}-03 & Nm \\
Negative Coulomb Friction & 1.15\text{E}-03 & 2.40\text{E}-03 & Nm \\
Motor Diameter & 40 & 30 & mm \\
Actuator Mass & 1.6680 & 0.5885 & kg\\
\hline
\end{tabular}
}%
\vspace{1mm}
\begin{minipage}{\columnwidth}
\footnotesize
\emph{Note:} Joint 1 and Joint 2 correspond to Actuator 2 and Actuator 3 from \cite{TUDOR-Online-Datasheet}
\end{minipage}
\end{table}

\subsection{Model 1: Ridge Regression with Gaussian Mixture Regressed Error}
\label{sec:Model_1}

Model 1 is proposed as a black-box approach that combines a parametric kinematics-based torque model with a non-parametric residual model. The parametric component is formulated as a linear regression problem by constructing a feature matrix from measured joint positions, velocities, and accelerations, as follows
\begin{equation}
\label{eq:Ridge_Model}
\boldsymbol{\tau}
=
\boldsymbol{\Phi}_{\text{kin}}(q,\dot{q},\ddot{q})\,\boldsymbol{w}.
\end{equation}
Ridge regression is then employed to estimate the corresponding weight vector, providing a baseline torque prediction on kinematic information. 

As with the other models considered in this study, the resulting torque residuals, defined as the difference between measured and predicted torque, are used to train a GMM. Residual torque predictions are then obtained via GMR and added to the parametric prediction to improve overall torque estimation accuracy. Diagonal covariance regularization is included in the GMM to improve numerical stability. 

\subsection{Model 2: Dynamics Modeling with Gaussian Mixture Regressed Error}
\label{sec:Model_2}
Model 2 employs the same 
GMR–based residual modeling framework described for Model 1 in Section \ref{sec:Model_1}, but couples it with a fully physics-based parametric model. The parametric torque component is obtained by evaluating the RBD formulation of the robot, as presented in Equation \ref{eq:expanded_tau}, where the grouped inertial and gravitational parameters $\alpha$ and $\beta$ are determined directly from the physical properties of the robot as provided in the documentation \cite{TUDOR-Online-Datasheet}.

Based on the schematic shown in Figure \ref{fig:TUDOR_schematic}, the primary actuator-related parameters governing the robot joints are summarized in Table \ref{tab:true_joint_params}. While link parameters are presented in Table \ref{tab:true_link_params}.

\begin{table}[H]
\centering
\caption{True Link Parameters from \cite{TUDOR-Online-Datasheet}}
\label{tab:true_link_params}
\small
\begin{tabular}{lccc}
\hline
\textbf{Parameter} & \textbf{Link 1} & \textbf{Link 2} & \textbf{Units} \\
\hline
Length & 430 & 410 & mm \\
Width & 15 & 15 & mm \\
Height & 4 & 4 & mm \\
Density & 7.8 & 7.8 & $10^{3}$ kg/m$^{3}$ \\
\hline
\end{tabular}
\end{table}

To determine the true dynamic parameters of the effective 2-DoF robotic arm, several simplifying assumptions were made. Using Table \ref{tab:true_joint_params}, for each link $i$, the mass $m_i$, center of mass (COM) location $l_{c,i}$, and Mass Moment of Inertia (MMoI) about the combined link–actuator COM $I_{c,i}$ are defined.

Given the physical properties of the links listed in Table \ref{tab:true_link_params}, the MMoI $I_{l,i}$, was first calculated about the COM of each link $i$, independently of the actuators using 
\begin{equation}
\label{eq:rod-MMoI}
    I_{l,i} = \frac{1}{12}m_{l,i}(h_{l,i}^2+l_{l,i}^2),
\end{equation}
where for isolated link $i$, $m_{l,i}$ is the mass, $h_{l,i}$ is the height, and $l_{l,i}$ is the length. The COM of each individual link was assumed to be at the midpoint of its length.

The first link and the second actuator were treated as a coupled rigid body. As a result, the effective mass of the first link $m_1$ was defined as the sum of the masses of link 1 and actuator 2. Given this mass offset, the effective COM location along link 1, $l_{c,1}$ was determined as
\begin{equation}
\label{eq:COM_Correction}
    l_{c,1} = \frac{(x_{l,1})(m_{l,1}) +(x_{a,2})(m_{a,2})}{m_{l,1}+m_{a,2}},
\end{equation}
where $x_{l,1}$ is the COM of link 1 on its own, $x_{a,2}$ is the COM of actuator 2 (assumed to be at the distal end of link 1), $m_{l,1}$ is the mass of link 1, and $m_{a,2}$ is the mass of actuator 2.

The MMoI was then updated for the coupled link 1, actuator 2 body. Assuming actuator 2 has cylindrical geometry,  the MMoI for actuator 2 was determined as
\begin{equation}
\label{eq:MMoI_Cylinder}
    I_{a,2} = \frac{1}{2}m_{a,2}r_{a,2}^2,
\end{equation}
where $r_{a,2}$ is the radius of actuator 2. The individual MMoIs of link 1 and actuator 2 were then offset to the new COM using parallel axis theorem to determine the final effective MMoI, $I_{c,1}$ for the coupled link 1 and actuator 2 body as 
\begin{align}
\label{eq:Parallel_Axis_MMoI}
    I_{c,1} =\Sigma( I_i + m_id_i^2),
\end{align}
where $I_i$ is the MMoI of each item about its COM, $m_i$ is the mass of each item, and $d_i$ is the distance between the item's COM and the new effective link COM. As no actuator is present at the end of link 2, the inertia parameter,$I_{c,2}$ corresponds solely to the inertia of link 2 such that $I_{c,2} = I_{l,2}$, thus defining all values required for $\alpha$ and $\beta$ in Equations \ref{eq:Inertia_Matrix} to \ref{eq:Gravity_Vector}.

To simplify the model, the viscous ($f_{v}$) and coulomb ($f_{c}$) frictions from Table \ref{tab:true_joint_params} were averaged across positive and negative directions of rotation, yielding a single viscous and Coulomb friction coefficient for each joint, thus fully defining Equation \ref{eq:Friction_Vector}.

With these parameters defined, Equation \ref{eq:expanded_tau} is fully specified and can be evaluated for each measurement sample. As link flexibility is not represented in the rigid-body formulation, substantial torque residuals are expected for the robot. These residuals are intentionally preserved and subsequently modeled using GMR to capture unmodeled elastic effects. The final parameter values used within the RBD model are presented in Table \ref{tab:RBD_model_params}.

\begin{table}[H]
\centering
\caption{Dynamic Parameters Used in RBD Model}
\label{tab:RBD_model_params}
\scriptsize
\renewcommand{\arraystretch}{0.8}
\setlength{\tabcolsep}{4pt}
\begin{tabular}{lclc}
\hline
Parameter & Value & Parameter & Value \\
\hline
$m_1$     & 1.869 kg                & $f_{v1}$  & 1.86E{-}05 N\,m\,s/rad \\
$m_2$     & 0.192 kg                & $f_{v2}$  & 2.45E{-}06 N\,m\,s/rad \\
$l_1$     & 0.430 m                 & $f_{c1}$  & 2.25E{-}03 N\,m \\
$lc_1$    & 0.407 m                 & $f_{c2}$  & 1.95E{-}03 N\,m \\
$lc_2$    & 0.205 m                 & $I_{c,2}$ & 2.69E{-}03 kg\,m$^2$ \\
$I_{c,1}$ & 3.10E{-}03 kg\,m$^2$  & & \\
\hline
\end{tabular}
\end{table}

Lastly, it should be noted that the torque constant and gear ratio values for each actuator are also provided in Table \ref{tab:true_joint_params}, as the dataset does not provide joint torque measurements directly, but instead reports motor current values. Joint torques were therefore computed as the product of motor current, torque constant, and gear ratio for each actuator.

\subsection{Model 3: Physics-Based Regression with Gaussian Mixture Regressed Error}
\label{sec:Model_3}

As Model 1 relies on a non-interpretable kinematics-based regression and Model 2 on a fully predefined physics-based parameter set, Model 3 is introduced as a gray-box approach adapted from the framework proposed by \cite{8978477}. This model combines a physics-informed parametric structure with data-driven parameter identification. Specifically, linear regression is used to estimate the inertial and gravitational parameters $\boldsymbol{\alpha}$ and $\boldsymbol{\beta}$, in addition to the frictional parameters, within the rigid-body dynamics formulation presented in Equation \ref{eq:expanded_tau}. In doing so, the parametric component is inferred directly from measured data while preserving the physical structure used in Model 2.

Unlike Model 2, which relies on explicit physical assumptions and parameter calculations, Model 3 estimates the dynamic parameters directly from data using regression, thereby avoiding manual computation of masses, centers of mass locations, and moments of inertia. The resulting parametric model remains physically interpretable and structurally consistent with rigid-body dynamics.

To enable parameter estimation via linear regression, Equation \ref{eq:expanded_tau} is recast in a linear-in-parameters form,
\begin{equation}
\label{eq:Parameter_Lin_regression}
\tau = Y(q,\dot{q},\ddot{q})\,\boldsymbol{\pi},
\end{equation}
where $Y(q,\dot{q},\ddot{q})$ denotes the regressor matrix and $\boldsymbol{\pi}$ is the vector of unknown dynamic parameters. For compact presentation, the transpose of the regressor matrix is shown,
\begin{equation}
\scriptsize
\mathbf{Y}(q,\dot q,\ddot q)=
\begin{bmatrix}
\ddot q_1 & 0 \\
2c_2\ddot q_1 + c_2\ddot q_2 - 2s_2\dot q_1\dot q_2 - s_2\dot q_2^2 &
c_2\ddot q_1 + s_2\dot q_1^2 \\
\ddot q_2 & \ddot q_1+\ddot q_2 \\
gc_1 & 0 \\
g(c_1+c_{12}) & gc_{12} \\
\dot q_1 & 0 \\
0 & \dot q_2 \\
\operatorname{sgn}(\dot q_1) & 0 \\
0 & \operatorname{sgn}(\dot q_2)
\end{bmatrix}^{\top}.
\end{equation}

As with the regressor matrix, for compact presentation, the parameter vector is also shown in transposed form:
\begin{equation}
\label{eq:pi_vector}
\boldsymbol{\pi} =
\begin{bmatrix}
\alpha_1 ,
\alpha_2 ,
\alpha_3 ,
\beta_1 ,
\beta_2 ,
f_{v1} ,
f_{v2} ,
f_{c1} ,
f_{c2}
\end{bmatrix}^{\top}.
\end{equation}

Using least-squares regression, a best-fit estimate of the parameter vector, denoted $\hat{\boldsymbol{\pi}}$, is obtained from training data. Similar to Model 2, Model 3 subsequently employs Gaussian Mixture Regression to model the residual torque error, which is added to the parametric prediction to obtain the final joint torque estimate.

\subsection{Offline Training and Online Adaptation}

The offline–online learning workflow for Models 1 to 3 (Sections \ref{sec:Model_1}–\ref{sec:Model_3}) consists of an initial offline training phase followed by an online adaptation phase in which data are processed sequentially. All models are first initialized using a training split of the selected subset of the robot arm dataset, after which predictions are generated for the remaining samples while model performance and residuals are evaluated against measured torque values.

In contrast to \cite{8978477}, where a dedicated excitation phase based on modified Fourier trajectories was used to ensure persistently exciting training data, the dataset used does not provide a separate training sequence with structured excitation. Moreover, the dataset contains extended intervals in which only a single joint is excited. To ensure sufficient excitation diversity during offline identification, the available data are shuffled prior to partitioning into training and streaming subsets. Empirical evaluation across multiple runs indicated that allocating approximately \TrainPercent of the dataset for offline training, followed by \TestPercent for online processing, provided a reliable balance between parameter convergence and generalization performance. 

Gaussian mixture residual models are trained during the offline phase using the residual torque obtained from each parametric model. To prevent numerical instability arising from near-singular covariance estimates, a covariance regularization constant of \CovRegulizer is applied. This value was found to stabilize training while having minimal influence on the learned residual structure.

During the online phase, samples are processed sequentially in time order. For Model 1, each new sample is appended to the existing training set and the ridge model is refit in batch using the expanded dataset. For Model 3, the physics-based model with estimated parameters $\hat{\pi}$, new samples are similarly appended to the accumulated regressor system and the parameter vector is re-estimated by re-solving a bounded least-squares problem at each step. In contrast, Model 2, which uses the known physical parameter vector $\pi_{True}$, remains fixed throughout the online phase and is not updated.

To maintain consistency between the residual Gaussian mixture model and the evolving parametric component in Models 1 and 3, the mean-only consistency transform proposed in \cite{8978477} is employed. When the parametric prediction changes as a result of online re-estimation, the Gaussian mixture output means are shifted by the corresponding change in predicted torque such that the residual model remains consistent with the updated parametric model. The covariance structure of the residual model is left unchanged.

\section{Results and Discussion}

\begin{figure}[t]
    \centering

    \begin{subfigure}[b]{\columnwidth}
        \centering
        \includegraphics[width=\linewidth]{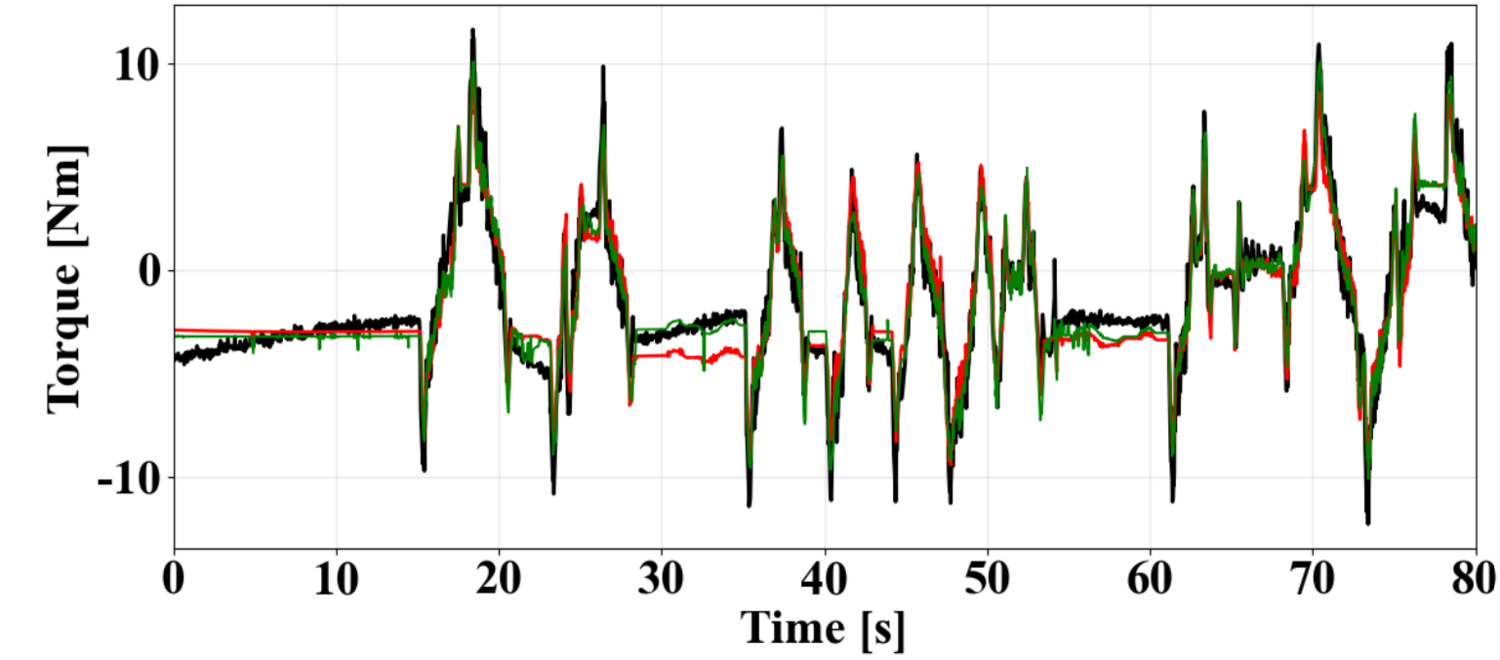}
        \caption{Model 1: Ridge Regression}
        \label{fig:J1T1}
    \end{subfigure}

    \vspace{0.5em}

    \begin{subfigure}[b]{\columnwidth}
        \centering
        \includegraphics[width=\linewidth]{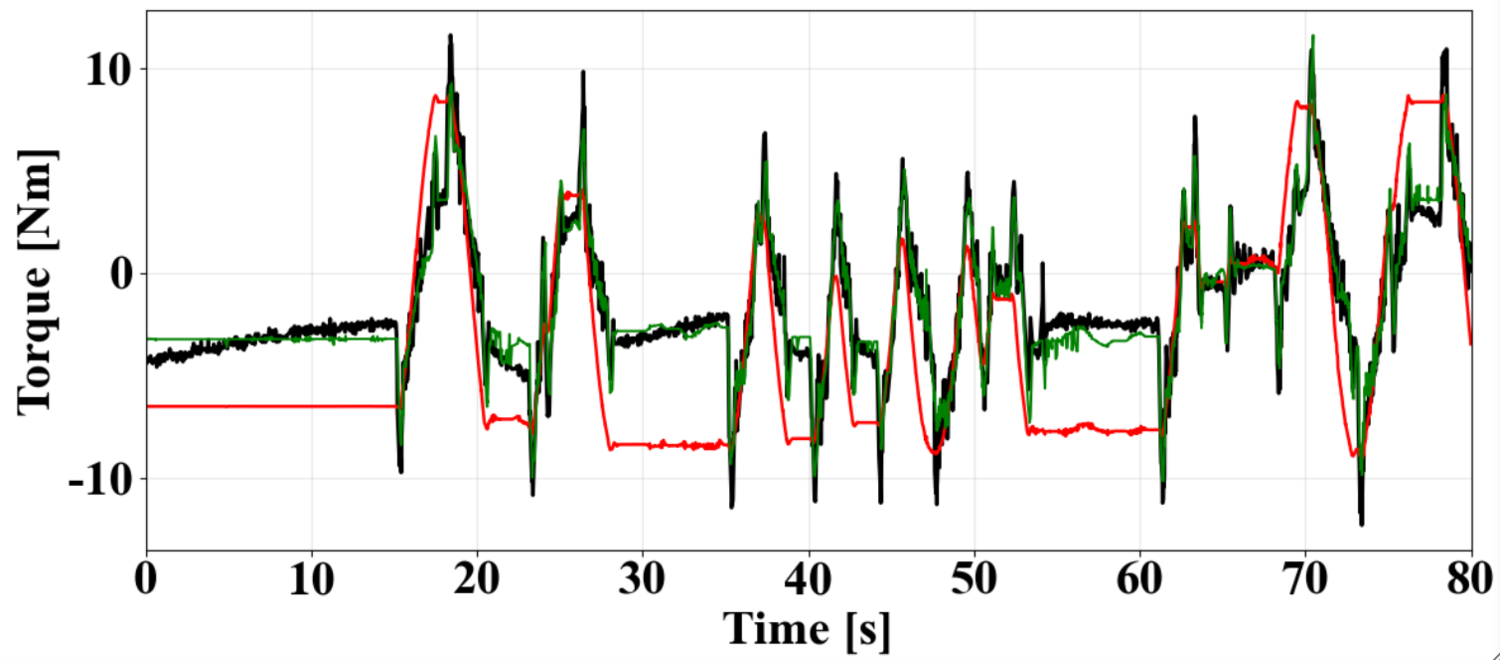}
        \caption{Model 2: Dynamics Modeling}
        \label{fig:J1T2}
    \end{subfigure}

    \vspace{0.5em}

    \begin{subfigure}[b]{\columnwidth}
        \centering
        \includegraphics[width=\linewidth]{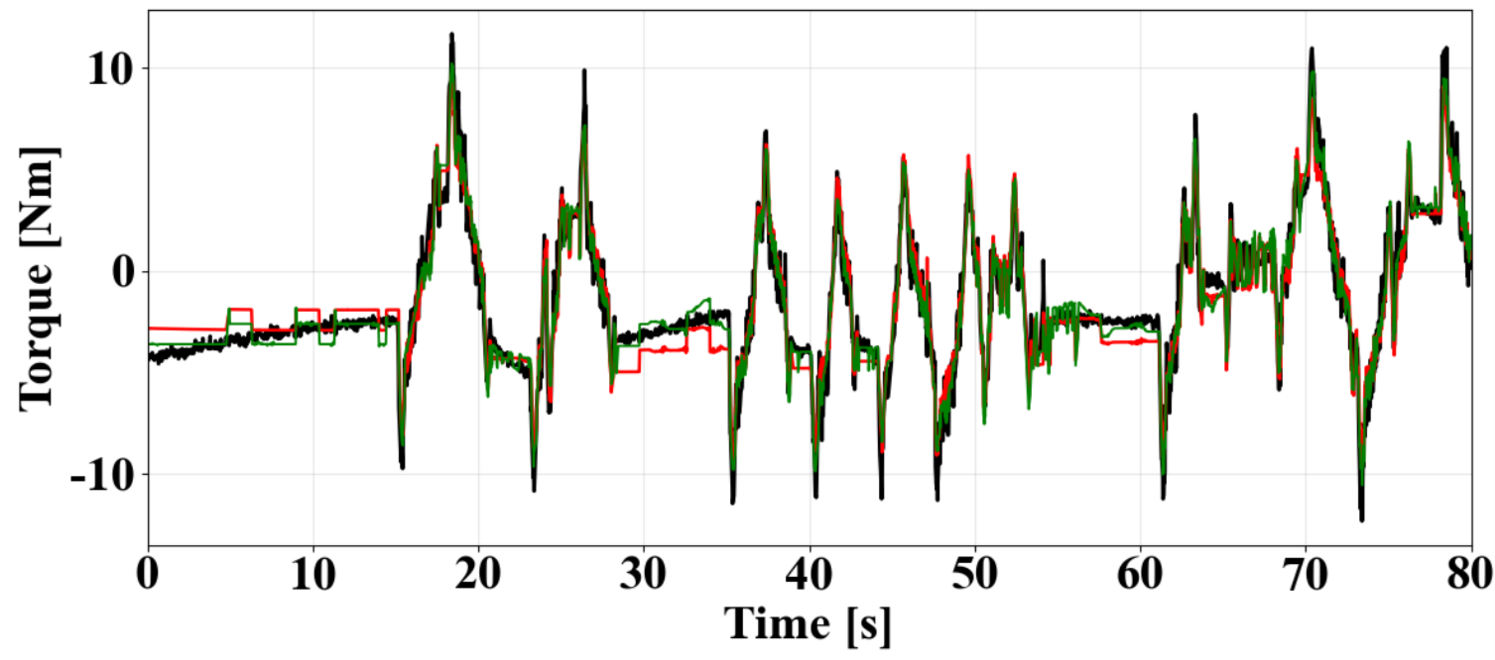}
        \caption{Model 3: Physics-Based Regression}
        \label{fig:J1T3}
    \end{subfigure}

    \caption{Joint 1 torque prediction results for three modeling approaches. The black curve plots ground-truth values, red is the parametric prediction of each model and green is the same prediction corrected with GMR.}
    \label{fig:J1_comparison}
\end{figure}

Predicted torque outputs are plotted with respect to measured torque values across the time domain. The torque estimates for all three models for joint 1 are shown in Figure \ref{fig:J1_comparison}. Red lines represent the parametric components corresponding to Models~1–3, representing a black-box data-driven approach, governed by Ridge regression; a white-box physics-based approach, governed by true physics parameters $\boldsymbol{\pi}_{\text{true}}$; and a gray-box physics-informed approach, governed by least-square determined parameters $\hat{\boldsymbol{\pi}}$ respectively. Green lines indicate model performance when GMR is used to estimate non-parametric residual dynamics. Black lines represent the measured joint torque values.

\begin{figure}[!t]
    \centering

    \begin{subfigure}[b]{\columnwidth}
        \centering
        \includegraphics[width=\linewidth]{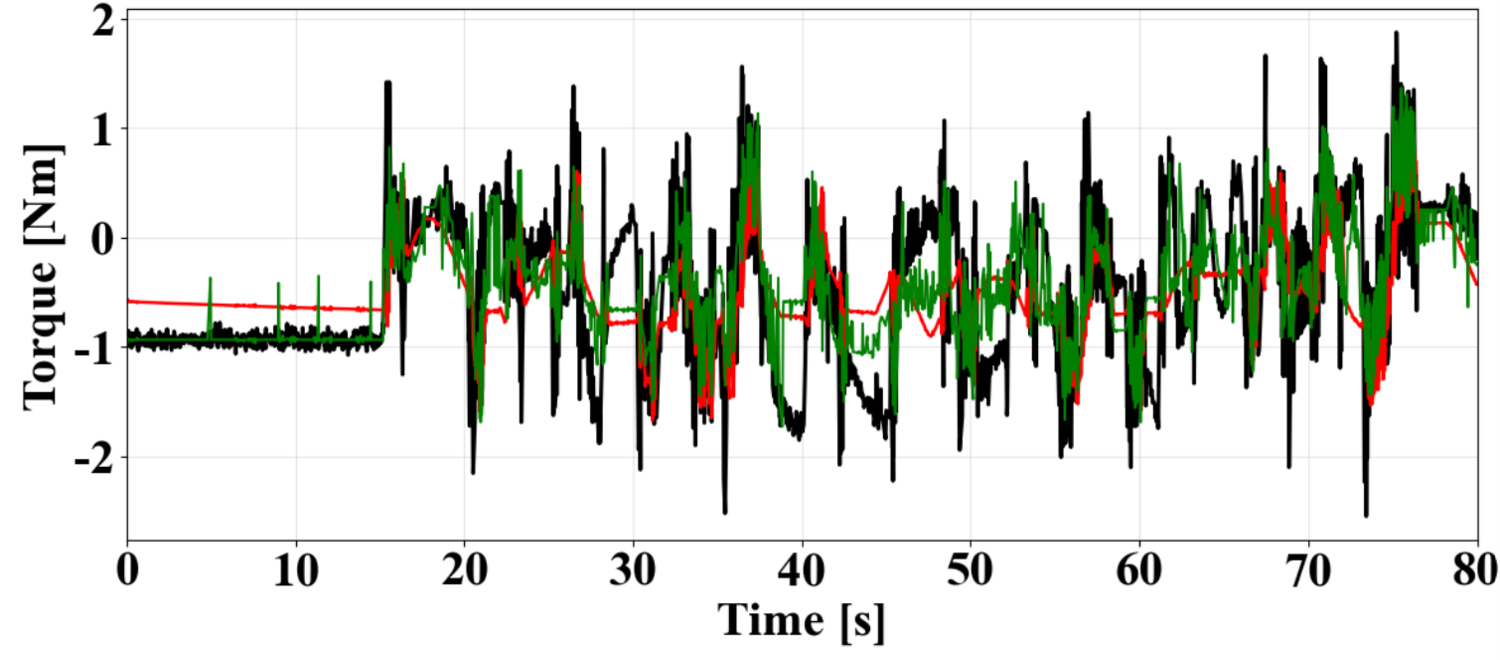}
        \caption{Model 1: Ridge Regression, joint 2}
        \label{fig:J2T1}
    \end{subfigure}

    \vspace{0.5em}

    \begin{subfigure}[b]{\columnwidth}
        \centering
        \includegraphics[width=\linewidth]{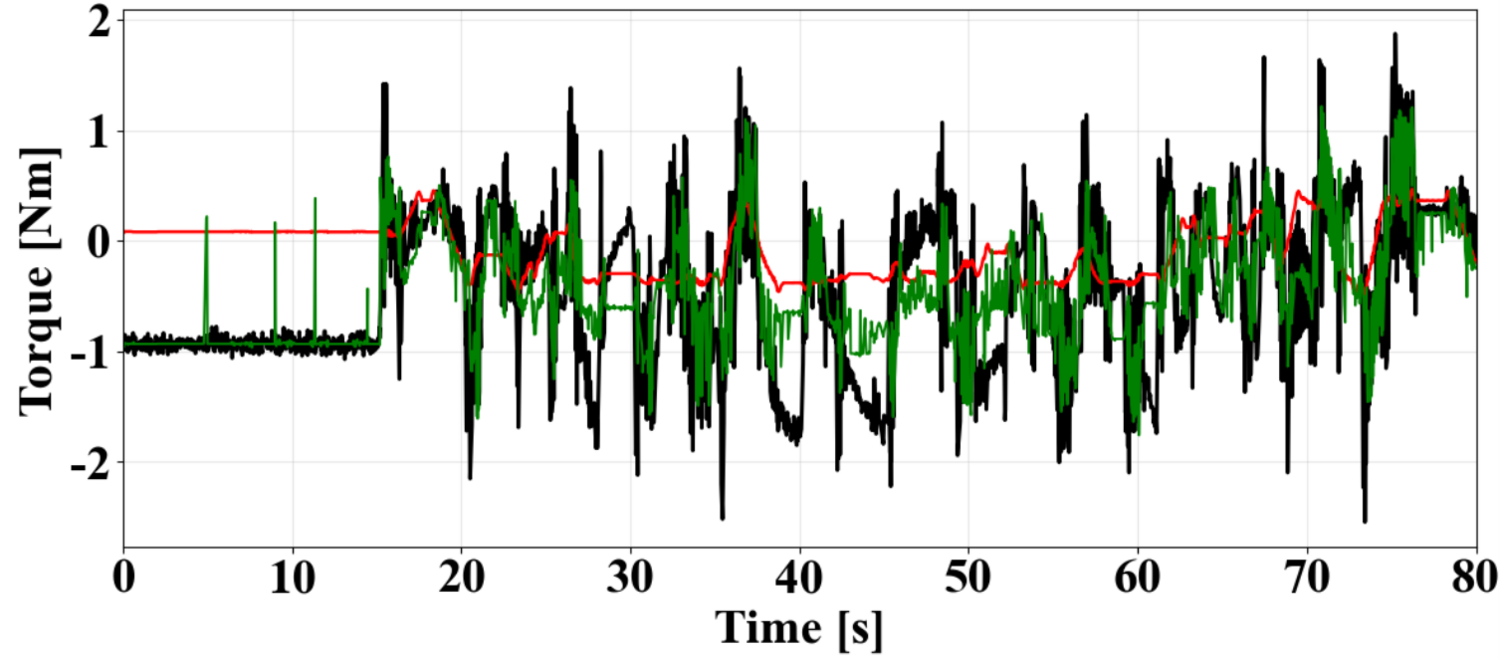}
        \caption{Model 2: Physics Formulation, joint 2}
        \label{fig:J2T2}
    \end{subfigure}

    \vspace{0.5em}

    \begin{subfigure}[b]{\columnwidth}
        \centering
        \includegraphics[width=\linewidth]{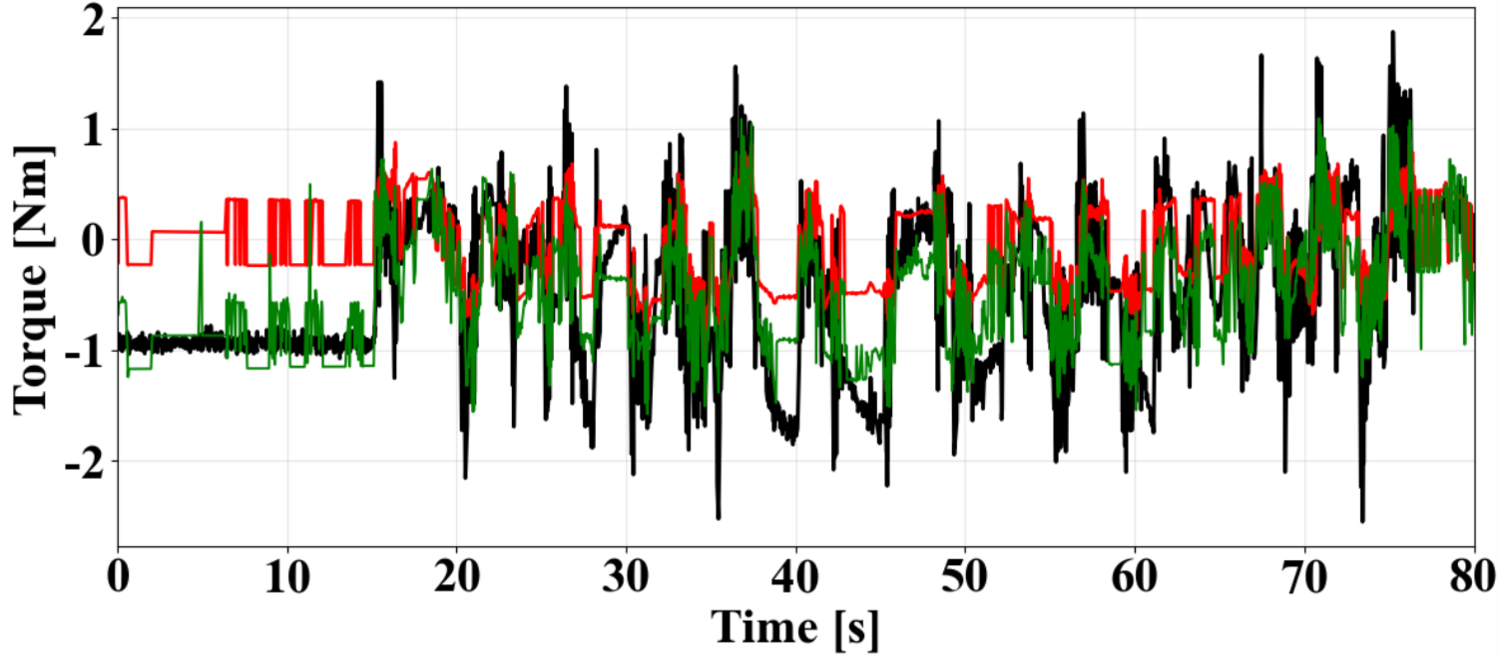}
        \caption{Model 3: Physics-Based Regression, joint 2}
        \label{fig:J2T3}
    \end{subfigure}

    \caption{Torque prediction results for joint 2 across the three modeling approaches.The black curve plots ground-truth values, red is the parametric prediction of each model and green is the same prediction corrected with GMR.}
    \label{fig:J2_comparison}
\end{figure}

Figures \ref{fig:J1_comparison} and \ref{fig:J2_comparison} illustrate torque deviations from the measured ground truth values. Additionally, the normalized Mean-Squared Error (nMSE) for each proposed model is calculated in Table \ref{tab:nMSE_values}.

\begin{table}[!h]
\centering
\caption{nMSE Calculated Across Validation of Models}
\label{tab:nMSE_values}
\begin{tabular}{lcc}
\hline
\textbf{Model} & \textbf{J1 nMSE Error} & \textbf{J2 nMSE Error} \\
\hline
RR & 0.002713& 0.01786\\
RR + GMR & 0.001911& 0.01243\\
DM & 0.02326& 0.03240\\
DM + GMR & 0.01810& 0.01256\\
PBR & 0.002139& 0.02711\\
PBR + GMR & 0.001578& 0.01126\\
\hline
\end{tabular}
\end{table}

\begin{figure}[!t]
\centering

\begin{subfigure}[b]{\columnwidth}
    \centering
    \includegraphics[width=\linewidth]{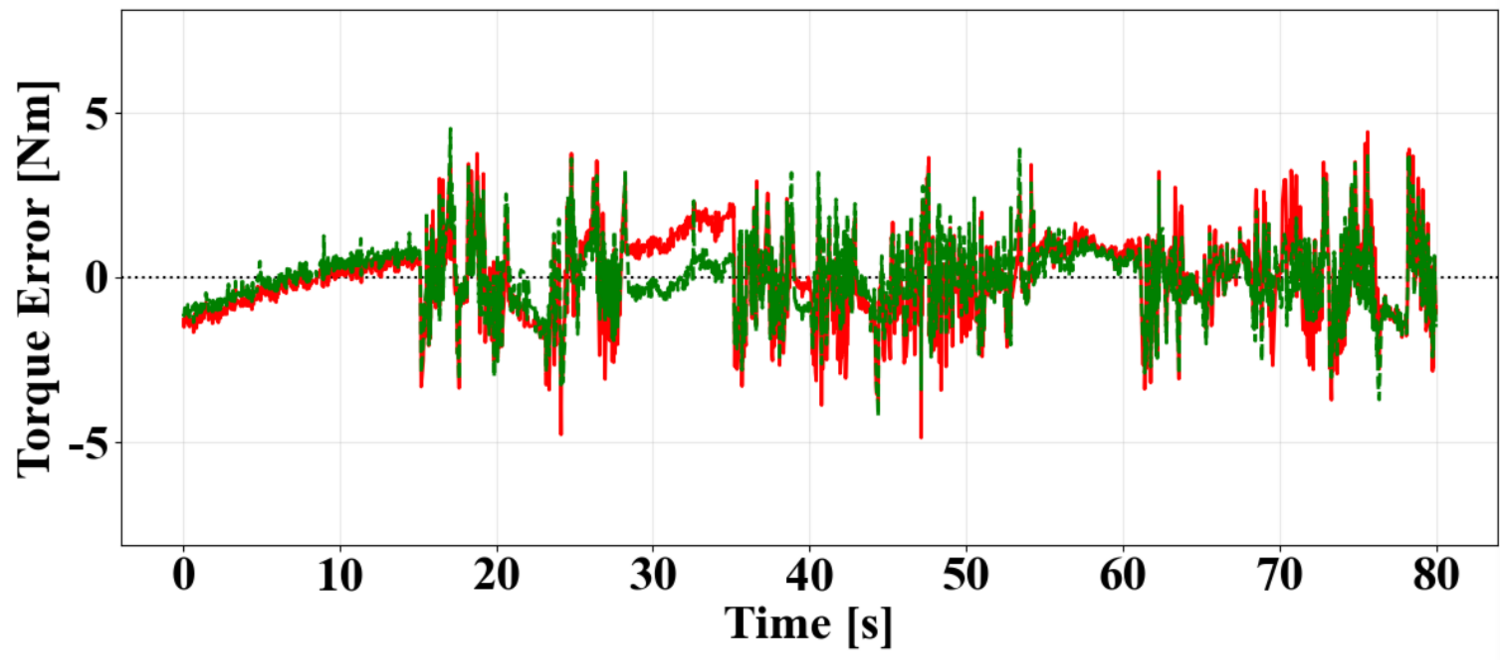}
    \caption{Model 1: Ridge Regression}
    \label{fig:J1E1}
\end{subfigure}

\vspace{0.25em}

\begin{subfigure}[b]{\columnwidth}
    \centering
    \includegraphics[width=\linewidth]{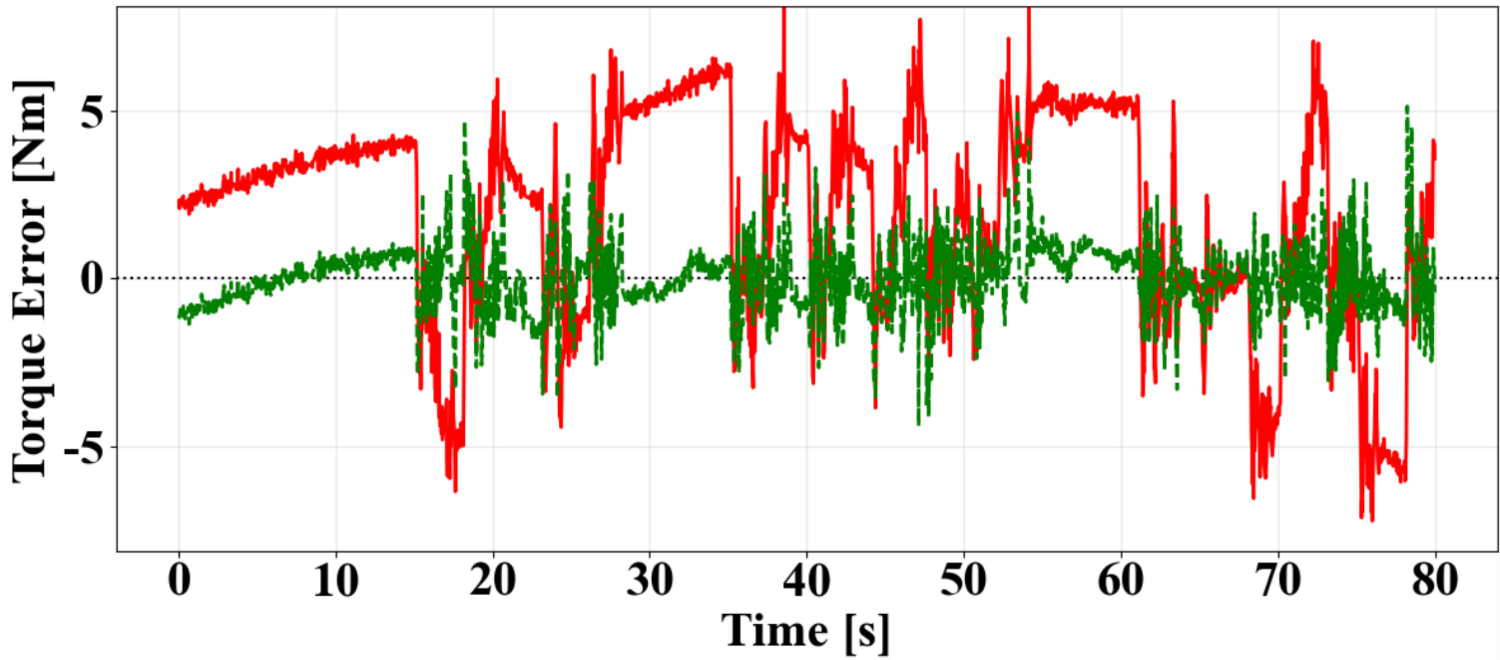}
    \caption{Model 2: Physics Formulation}
    \label{fig:J1E2}
\end{subfigure}

\vspace{0.25em}

\begin{subfigure}[b]{\columnwidth}
    \centering
    \includegraphics[width=\linewidth]{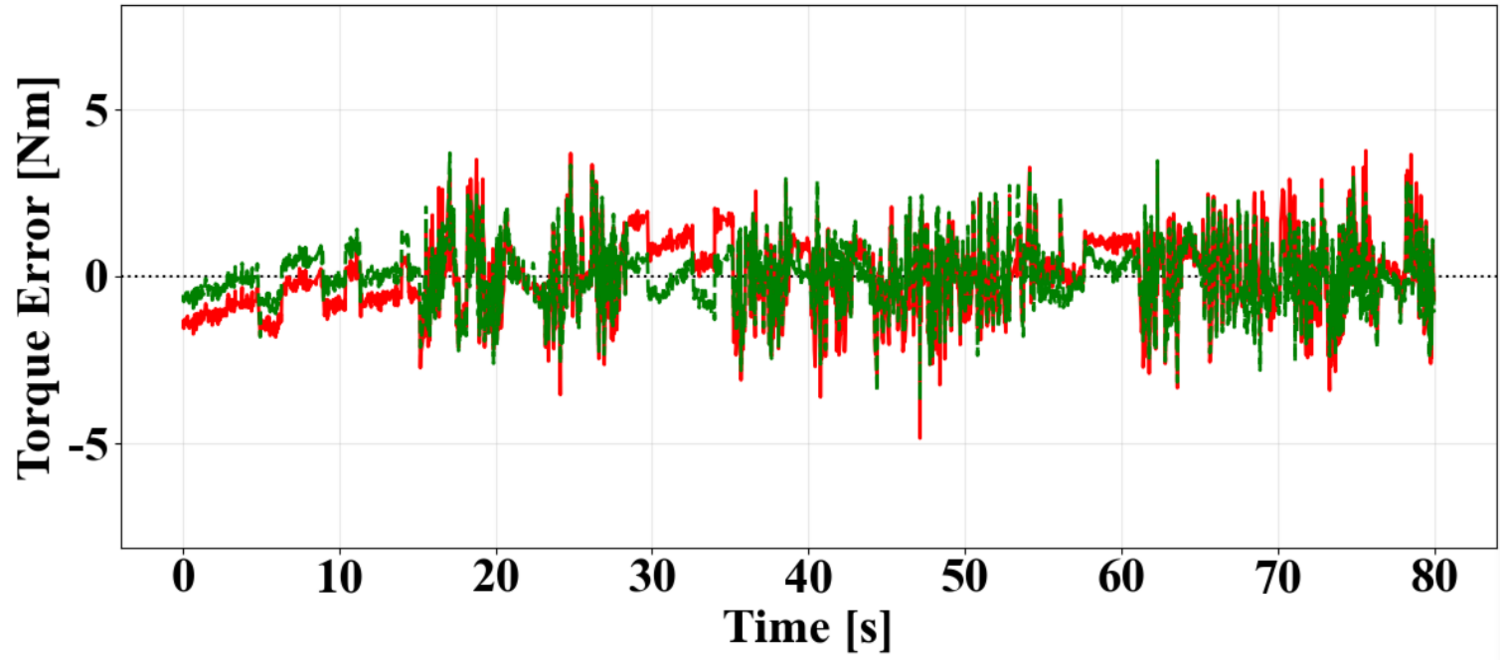}
    \caption{Model 3: Physics-Based Regression}
    \label{fig:J1E3}
\end{subfigure}

\caption{Torque error comparison for joint~1 across the three modeling approaches.
Red indicates the purely parametric prediction error, while green indicates the error after applying GMR.}
\label{fig:torque_error_j1}
\end{figure}

\begin{figure}[!t]
\centering

\begin{subfigure}[b]{\columnwidth}
    \centering
    \includegraphics[width=\linewidth]{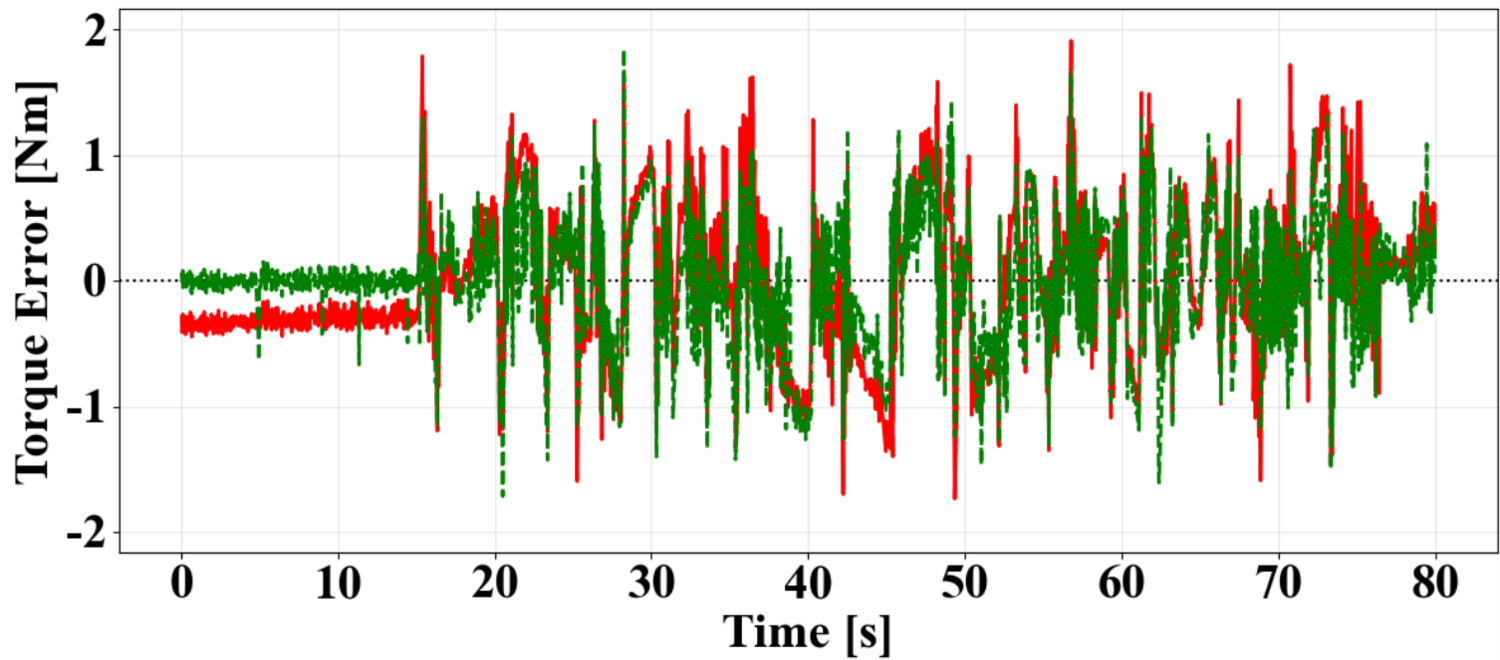}
    \caption{Model 1: Ridge Regression}
    \label{fig:J2E1}
\end{subfigure}

\vspace{0.25em}

\begin{subfigure}[b]{\columnwidth}
    \centering
    \includegraphics[width=\linewidth]{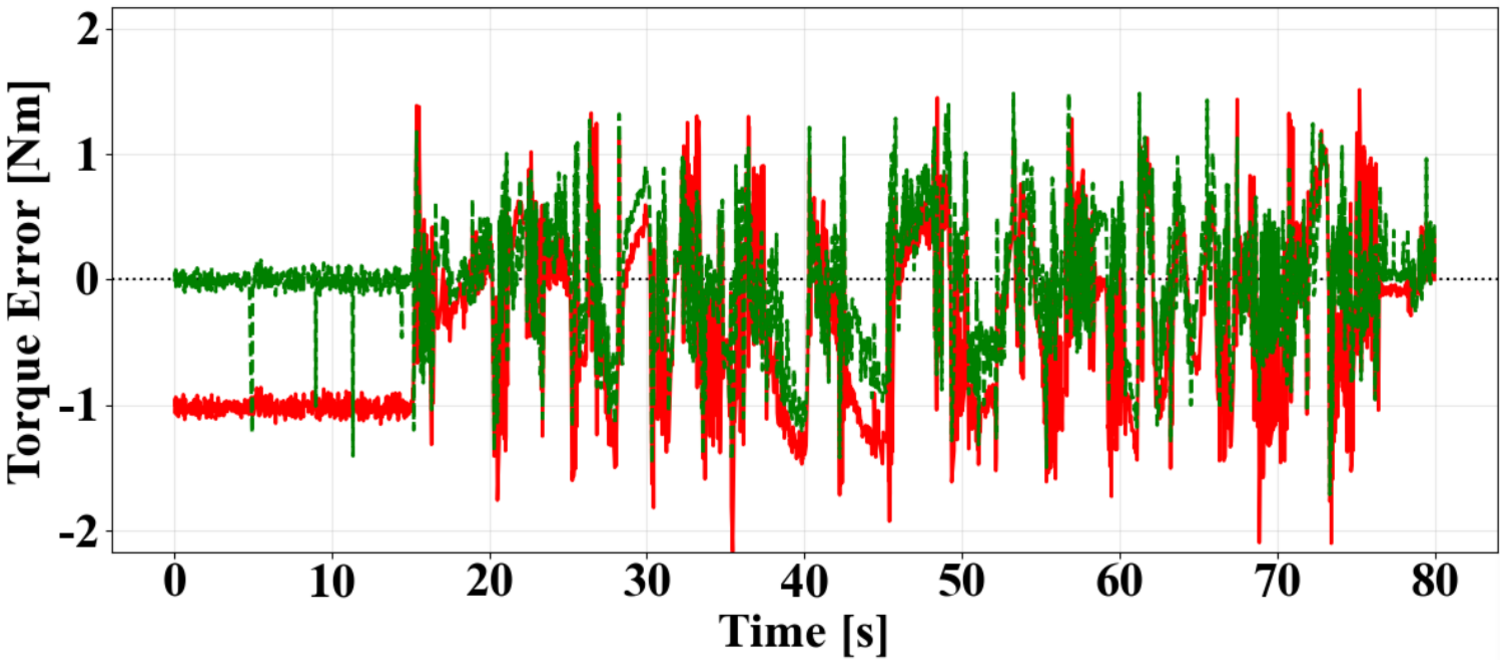}
    \caption{Model 2: Physics Formulation}
    \label{fig:J2E2}
\end{subfigure}

\vspace{0.25em}

\begin{subfigure}[b]{\columnwidth}
    \centering
    \includegraphics[width=\linewidth]{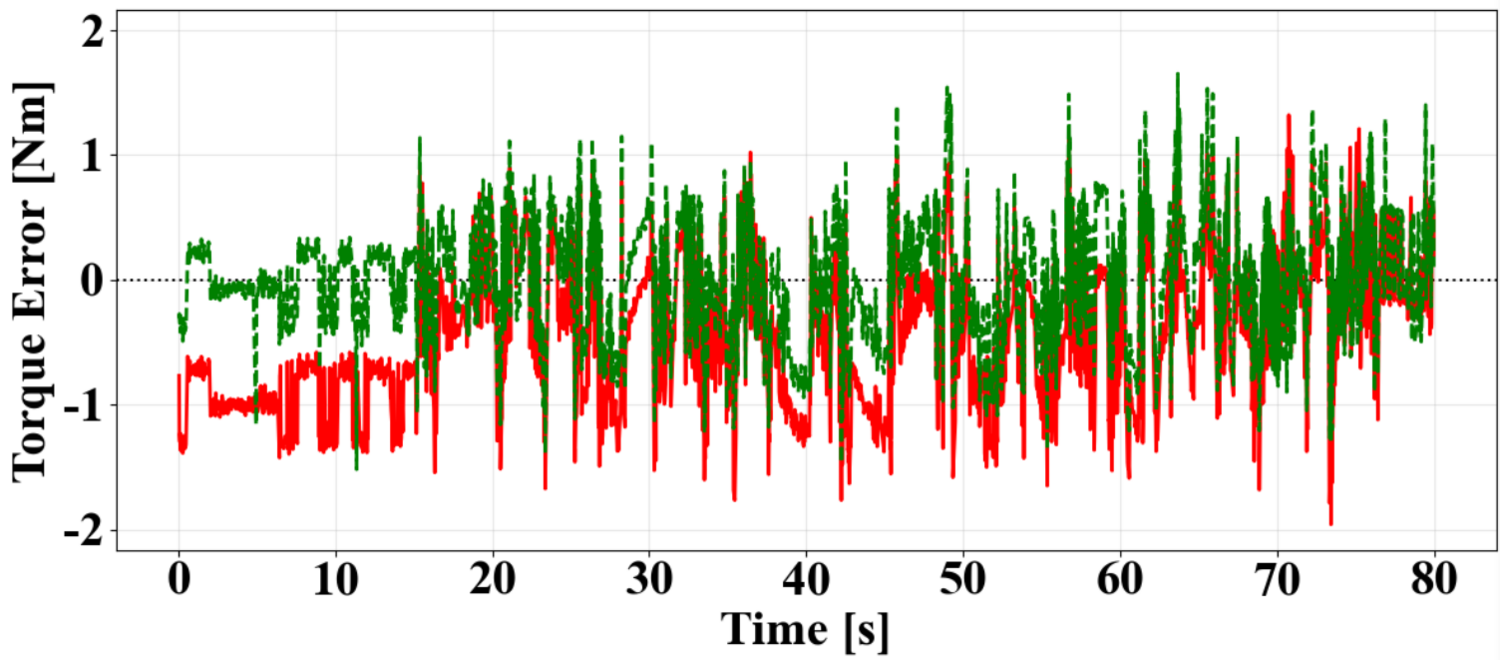}
    \caption{Model 3: Physics-Based Regression}
    \label{fig:J2E3}
\end{subfigure}

\caption{Torque error comparison for joint~2 across the three modeling approaches.
Red indicates the purely parametric prediction error, while green indicates the error after applying GMR.}
\label{fig:torque_error_j2}

\end{figure}

It is observed that with adequate training data, the Physics-Based Gaussian Mixture Regression model outperforms all other models in torque estimation accuracy across both joints. As the total normalized mean squared error is representative of all idling, actuating, and overshooting movements it is important to observe how each model performs on a per motion basis. Figure \ref{fig:torque_error_j1} and \ref{fig:torque_error_j2} illustrate the torque error across the 80 second motion trial.

From these figures, it is observed that Gaussian Mixture Regression is able to correct the majority of base model deviations from the ground truth in idle positions across all three models. The largest deviations from the ground truth occur during actuations which induce oscillatory motion. However, as these trends hold true across all models, the highest fitting error still occurs within physics informed dynamic modeling method. Figure \ref{fig:J1E2} illustrates that although the resulting formulation of Equation \ref{eq:General_Torque_Equation} possesses the most predictable parametric torque response of all three models, it lacks estimation accuracy when compared to measured torque values (see Section: \ref{sec:Model_2} for details). This is because the model neither accounts for vibrational fluctuations nor the large disturbance forces induced by the flexible joints of the robot. However, through implementing GMR this error is heavily reduced to the point in which it performs with accuracy that is comparable to that of the other models, as shown in Table \ref{tab:nMSE_values}. It should be noted however that without GMR, the predictions of Model 2 are significantly worse as compared to those from learning-based methods of Models 1 and 3.

Figures \ref{fig:torque_error_j1} and \ref{fig:torque_error_j2} illustrate that the black-box Model 1 is very comparable to the least squares semi-parametric Model 3. Without the inherent understanding of the robot arm physics, this model is still capable of determining meaningful relations with minimal modeling effort compared to Models 2 and 3. Inspection of the fitted Ridge coefficients also revealed a comparatively large intercept term, particularly for joint~1. This is consistent with the baseline mismatch observed when using the purely physics-based dynamic model of Model~2, as visible in Figure~\ref{fig:J1T2}. Since Model~1 uses a linear kinematic feature map rather than structured rigid-body gravity and friction terms, its intercept provides a direct way to shift the prediction baseline. In Model~1, this offset is therefore captured as a constant predictive bias rather than as a physically interpretable rigid-body parameter. This further reinforces the role of Model~1 as a predictive black-box baseline rather than a physically interpretable dynamics model.

The gray-box model combines the adaptability of black-box learning with the structural constraints of a 2-DOF robotic arm allowing for the lowest baseline model error while ensuring that GMR correction effects remain significant.

For interpreting the gray-box model regression, the magnitudes of the fitted parameters presented in Equations \ref{eq:Inertia_Matrix} to \ref{eq:Friction_Vector} are analyzed in Table \ref{tab:least_square_param_compare}. The grouped parameters fall into three physical categories. The inertial parameters $\alpha_1$, $\alpha_2$, and $\alpha_3$ represent the total effective inertia at joint 1, the inertial coupling between the two joints, and the effective inertia at joint 2, respectively. The gravity parameters $\beta_1$ and $\beta_2$ capture the gravitational loading at joint 1 and joint 2, respectively. Finally, $f_{v1}$ and $f_{v2}$ represent viscous friction at each joint, while $f_{c1}$ and $f_{c2}$ represent Coulomb friction at each joint.

Within Table \ref{tab:least_square_param_compare}, it was observed that, in comparison to the true parameters of Model 2, the regression overestimated the inertial parameters of both linkages, underestimated the Coriolis effect, and heavily overestimated Coulomb and Viscous frictional forces. The overestimation of $\alpha_1$ and $\alpha_3$ suggests the model perceives the arm as more resistant to acceleration than the physical parameters indicate, while the underestimation of $\alpha_2$ implies the data-driven approach does not fully capture the inertial coupling between the two joints due to elastic effects. The underestimation of $\beta_1$ and $\beta_2$ indicates the regression attributes less gravitational loading to each joint than is physically present. Bending of the robotic arm’s linkages increases the actuating torque required for a desired angular acceleration by creating a delayed inertial response. This phenomenon is likely absorbed into the model's frictional terms, as it mainly occurs both during the starting and stopping conditions of the actuator, and scales proportionally to actuating speed, which is consistent with the observed overestimation of both $f_{v}$ and $f_{c}$ terms. However, Figures \ref{fig:J1T2} and \ref{fig:J2T2}, show that the regularized GMR succeeds in accounting for these modeling inconsistencies by grouping them with non-parametric errors which allows the predicted response to closely match with ground truth torque measurements.

To assess the numerical stability of the least-squares identification used in Model~3, the rigid-body dynamics regressor matrix $Y$ was examined on the training data. The regressor was found to be full column rank, and the corresponding condition number was 23.67, indicating that the least-squares problem was not strongly ill-conditioned. This suggests that the large deviation between $\hat{\boldsymbol{\pi}}$ and $\boldsymbol{\pi}_{\text{true}}$ is more likely attributable to rigid-body model mismatch and the absorption of flexible-link effects into effective surrogate parameters, rather than numerical degeneracy of the regression itself.

\begin{table}[!h]
\vspace{10pt}
\centering
\caption{Linear Least Squares Solution for $\hat{\boldsymbol{\pi}}$, used in the Training of Model 3, Compared with True RBD Parameters, used in Model 2}
\label{tab:least_square_param_compare}
\begin{tabular}{lccc}
\hline
\textbf{Parameter} & $\hat{\boldsymbol{\pi}}$ \textbf{Value} & $\boldsymbol{\pi}_{\text{True}}$ \textbf{Value} & \textbf{Units} \\
\hline
$\alpha_1$ & $7.6283\times10^{-1}$ & $3.6571\times10^{-1}$ & kg$\cdot$m$^2$ \\
$\alpha_2$ & $5.2071\times10^{-2}$ & $1.6914\times10^{-2}$ & kg$\cdot$m$^2$ \\
$\alpha_3$ & $4.0211\times10^{-2}$ & $1.0752\times10^{-2}$ & kg$\cdot$m$^2$ \\
$\beta_1$  & $3.6403$             & $8.2331$             & N$\cdot$m \\
$\beta_2$  & $3.6291\times10^{-1}$ & $3.8588\times10^{-1}$ & N$\cdot$m \\
$f_{v1}$   & $1.1537$             & $1.8600\times10^{-5}$ & N$\cdot$m$\cdot$s/rad \\
$f_{v2}$   & $9.7576\times10^{-2}$ & $2.4500\times10^{-6}$ & N$\cdot$m$\cdot$s/rad \\
$f_{c1}$   & $9.9977\times10^{-1}$ & $2.2500\times10^{-3}$ & N$\cdot$m \\
$f_{c2}$   & $2.9952\times10^{-1}$ & $1.9500\times10^{-3}$ & N$\cdot$m \\
\hline
\end{tabular}
\vspace{-4pt}
\end{table}

\begin{figure}[ht]
\begin{center}
\includegraphics[width=\linewidth]{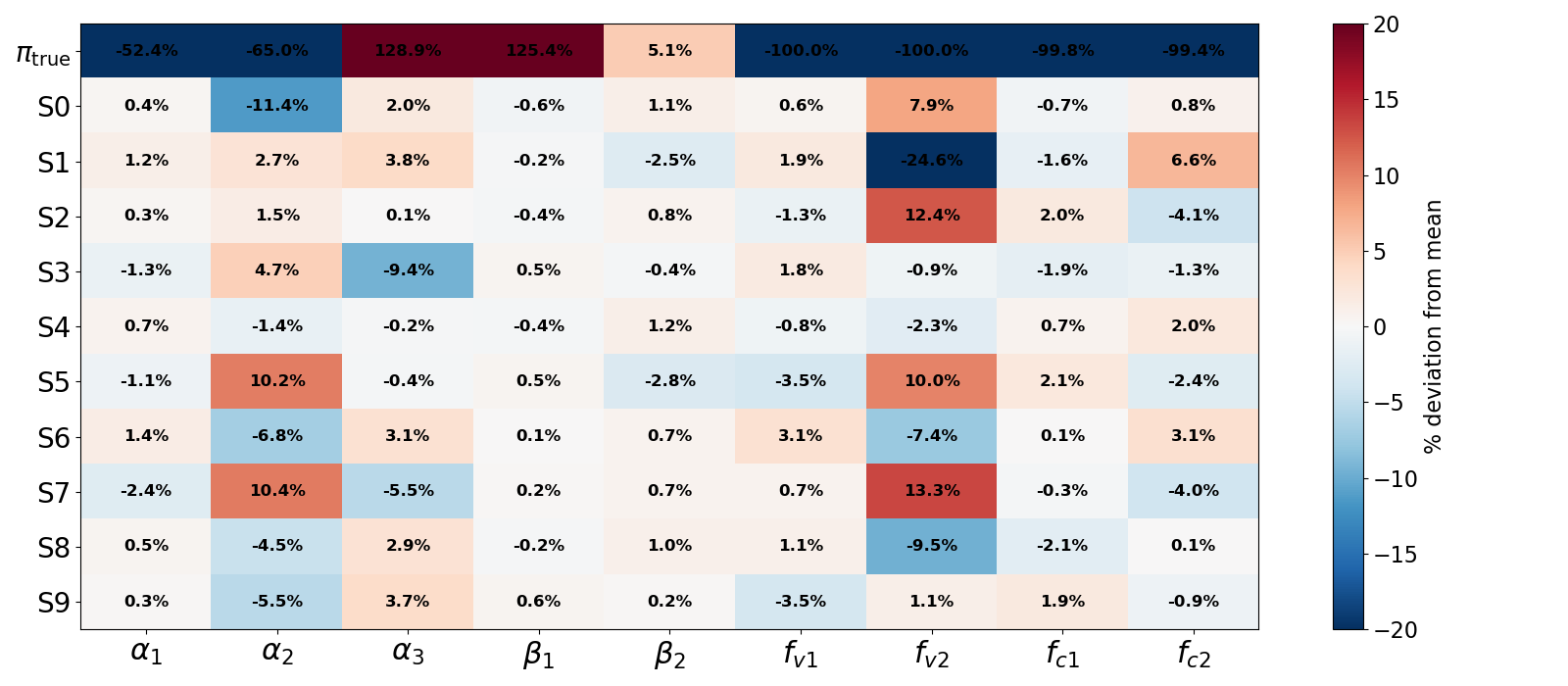}
\caption{Grouped Parameters Regressed across 10 different Seed Values.}
\label{fig:seedmap}
\end{center}
\end{figure}

To confirm that these model parameters are converged within the provided data, Figure \ref{fig:seedmap} shows 10 different seed variations with the same train-test split. A heatmap is applied to illustrate how grouped parameters differ from the observed parameter mean across all 10 seeds. Furthermore, Table \ref{tab:torque_error_stats} reports the mean and variance of torque error across each seed. 

\begin{table}[!h]
\vspace{-4pt}
\centering
\caption{Torque Error Statistics Across 10 Seeds}
\label{tab:torque_error_stats}
\begin{tabular}{lccccc}
\hline
\textbf{Method} & \textbf{Joint} & \textbf{Mean RMSE} & \textbf{Std RMSE} \\
\hline
Ridge & Joint 1 & $1.26026$ & $1.802\times10^{-2}$ \\
Ridge & Joint 2 & $0.58937$ & $6.18\times10^{-3}$ \\
RBD ($\pi_{\text{hat}}$) & Joint 1 & $1.11702$ & $2.005\times10^{-2}$ \\
RBD ($\pi_{\text{hat}}$) & Joint 2 & $0.71573$ & $5.18\times10^{-3}$ \\
\hline
\end{tabular}
\vspace{-4pt}
\end{table}

From Figure \ref{fig:seedmap}, it is observed that the majority of grouped parameters have steadily converged and show little variance in value across different seeds. However, certain variables such as $\alpha_2$, which denotes the inertial coupling effects between joints, and $f_{v2}$ which denotes the viscous friction effects within the second joint show deviations within $15\%$. This likely arises from the aforementioned bending effects and has little impact on predicted torque accuracy as reflected in RMSE values within Table \ref{tab:torque_error_stats} having standard deviations below $2\%$.

\begin{figure}[t]
\centering

\noindent\makebox[\linewidth]{%
\begin{minipage}[b]{0.32\linewidth}\centering\small\textbf{Model 1}\end{minipage}\hfill
\begin{minipage}[b]{0.32\linewidth}\centering\small\textbf{Model 2}\end{minipage}\hfill
\begin{minipage}[b]{0.32\linewidth}\centering\small\textbf{Model 3}\end{minipage}%
}

\vspace{0.3em}

\begin{subfigure}[b]{0.32\linewidth}
  \centering
  \includegraphics[width=\linewidth]{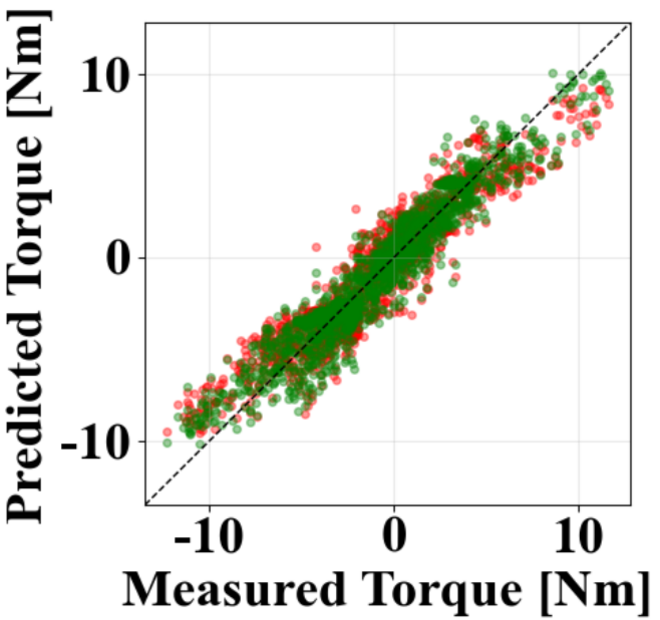}
\end{subfigure}\hfill
\begin{subfigure}[b]{0.32\linewidth}
  \centering
  \includegraphics[width=\linewidth]{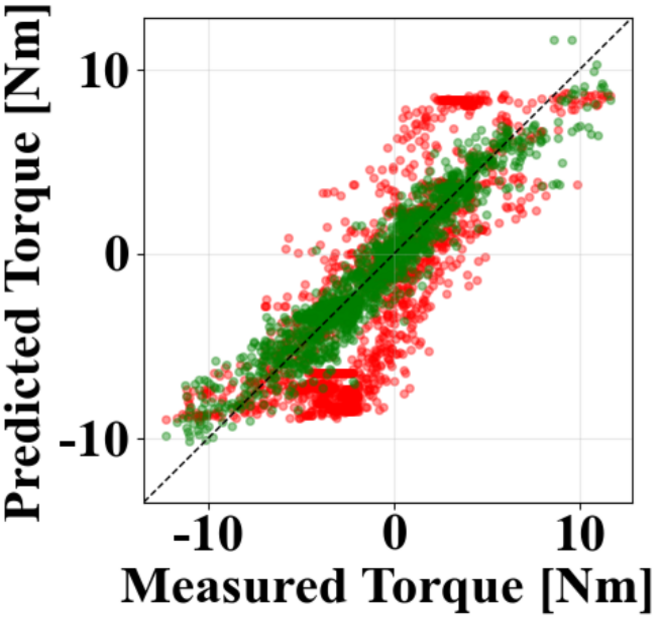}
\end{subfigure}\hfill
\begin{subfigure}[b]{0.32\linewidth}
  \centering
  \includegraphics[width=\linewidth]{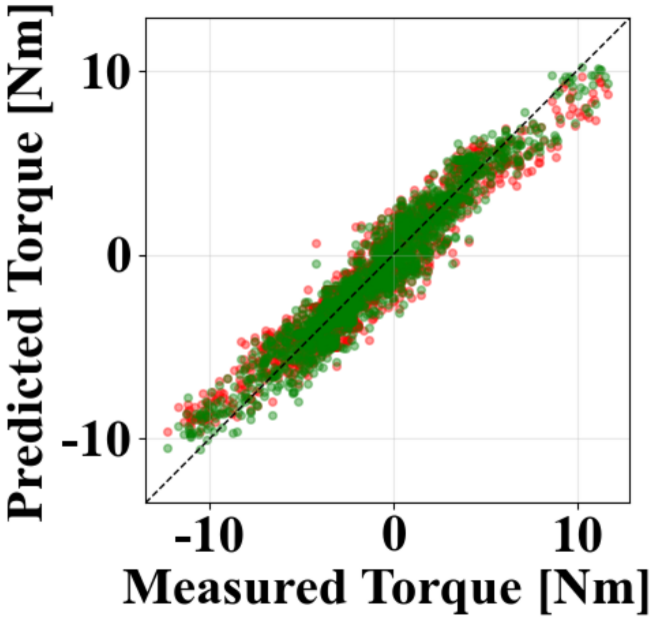}
\end{subfigure}

\vspace{0.4em}

\begin{subfigure}[b]{0.32\linewidth}
  \centering
  \includegraphics[width=\linewidth]{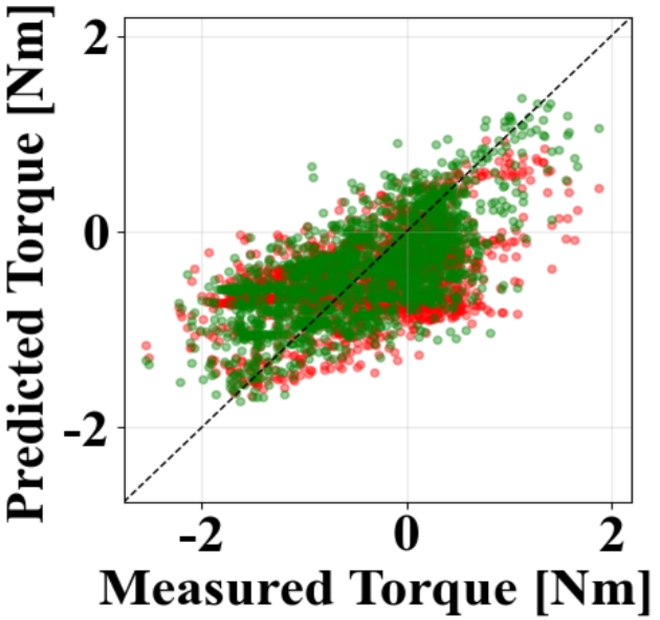}
\end{subfigure}\hfill
\begin{subfigure}[b]{0.32\linewidth}
  \centering
  \includegraphics[width=\linewidth]{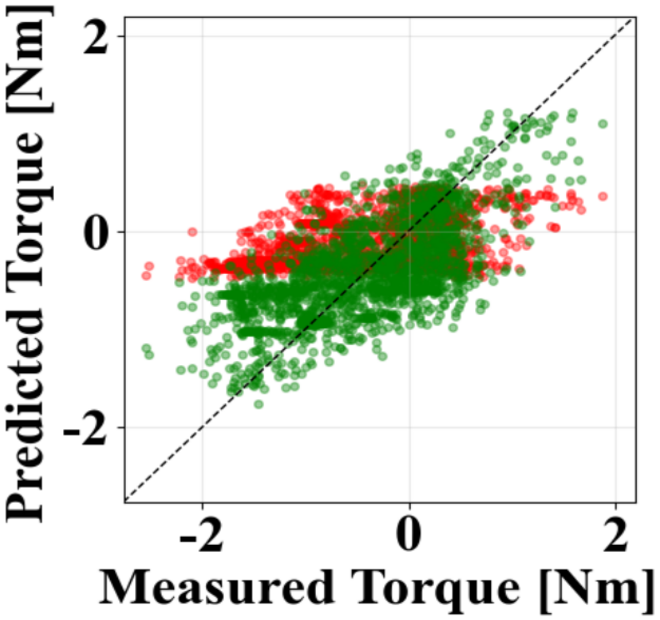}
\end{subfigure}\hfill
\begin{subfigure}[b]{0.32\linewidth}
  \centering
  \includegraphics[width=\linewidth]{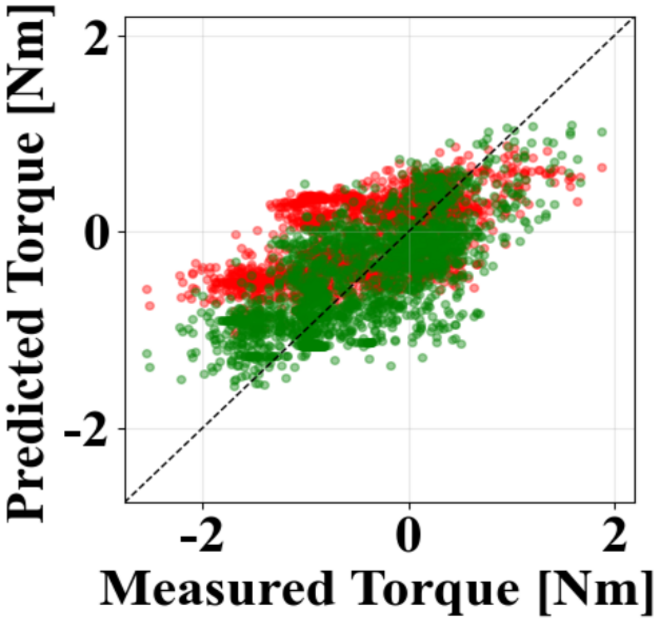}
\end{subfigure}

\caption{Parity plots for Joint~1 (top row) and Joint~2 (bottom row) for Model~1 (left column), Model~2 (center column) and Model~3 (right column).}
\label{fig:parity}
\end{figure}

Of all the parity plots developed in Figure \ref{fig:parity}, the physics-based Model 2 had the worst performance without GMR. This is likely also due to the delayed inertial response introduced by the flexible linkages amplifying the real torque output of the actuators. Model 2 makes a complete RBD assumption with joint and linkage parameters taken from the online robot arm specifications sheet \cite{TUDOR-Online-Datasheet}, leaving no opportunity to compensate for linkage flexibility. On the other hand, the physics-based regression model possessed the best alignment between torque estimations and measurements as it was able to capture the effects of bending in robot kinematics through compensation of predicted terms, especially overestimating the friction terms, in its model. The parity plot displayed in Figure~\ref{fig:parity} also illustrates that after applying GMR, all model predictions align well with joint 1 torque measurements as plotted data points display a linear trend with a narrow spread. The significant overlap between green and red data points also indicates that due to the satisfactory performance of the parametric models, the GMM does not have to fit a significant amount of non-parametric errors. On the other hand, all models exhibit reduced performance at joint 2, as the uncertainty propagated from joint 1 is compounded by the uncertainty generated by linkage flexure in joint 2. 

\begin{figure}[!t]
\centering

\begin{subfigure}[b]{\columnwidth}
    \centering
    \includegraphics[width=\linewidth]{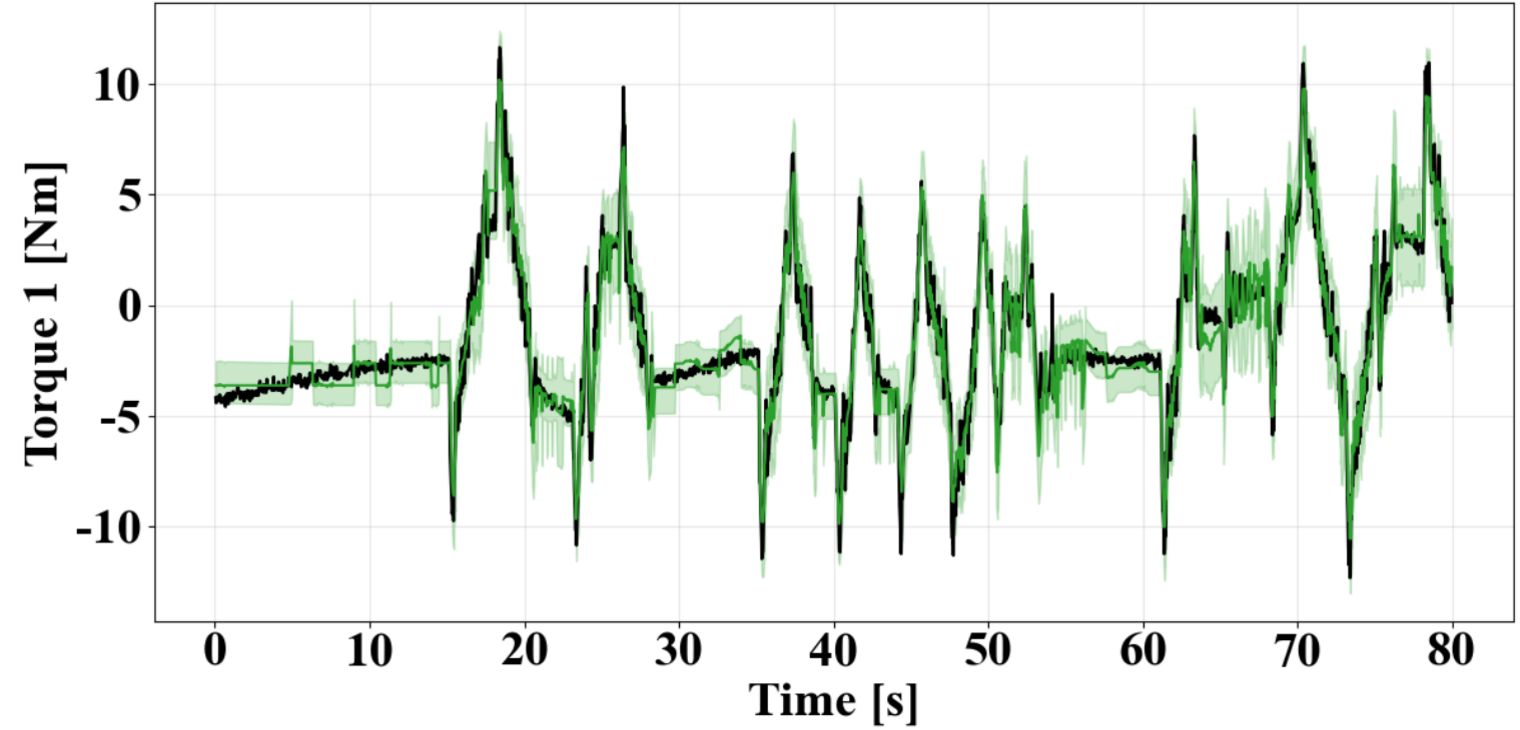}
    \caption{Actuator One}
    \label{fig:CI_J1}
\end{subfigure}

\vspace{0.5em}

\begin{subfigure}[b]{\columnwidth}
    \centering
    \includegraphics[width=\linewidth]{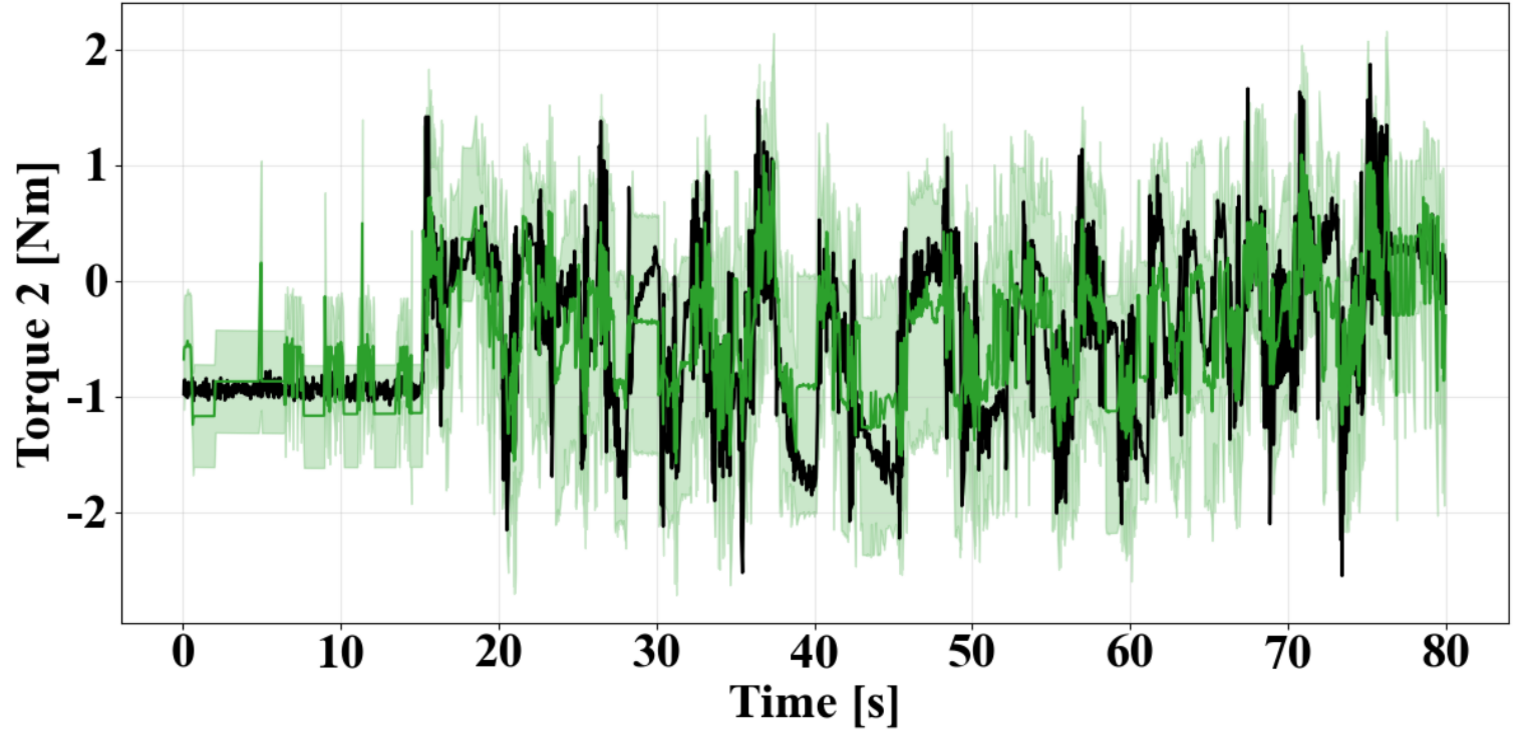}
    \caption{Actuator Two}
    \label{fig:CI_J2}
\end{subfigure}

\caption{Torque response with 95\% confidence intervals for Actuators One and Two.}
\label{fig:CI}
\end{figure}

From the previous results illustrating that the physics-based semi-parametric model provides the most accurate torque modeling along with the findings presented in \cite{8978477}, the model that should be applied in the online autonomous control of a robotic arm is the adapted gray-box model. Figures \ref{fig:CI_J1} and \ref{fig:CI_J2} depict the 95\% confidence intervals of this model's time response. Small confidence intervals across joint 1 indicate that the physics based regression performed in estimating the $\hat{\pi}$ matrix is true to the data, such that accurate predictions can be made. Although the confidence intervals across the joint 2 response are wider, their magnitudes are comparable to measurement fluctuations such that the predicted torque values are not completely random and the systems true state may still be accurately predicted.


\section{Conclusions}

In this work, a semi-parametric gray-box torque modeling framework was applied to a lightweight, flexible-link two-degree-of-freedom robotic arm. The approach combines parametric rigid-body dynamics models with non-parametric residual learning using Gaussian mixture regression, enabling improved torque prediction in the presence of structural flexibility and unmodeled dynamics. To ensure convergence stability in fitting, a regularization term of $\CovRegulizer$ is applied to the model.


Classical rigid-body dynamics formulations assume rigid links, fixed inertial parameters, and simplified friction models, and thus cannot explicitly account for elastic deformation, joint compliance, or load-dependent variations in inertia. The selected data set \cite{TUDOR-MERIt-2014} was taken from a robotic arm which was intentionally designed with a lightweight structure, making these effects significant relative to conventional rigid manipulators and leading to substantial prediction errors for purely physics-based models.

The proposed semi-parametric approach combines the interpretability and structure of rigid-body dynamics with the expressive power of data-driven learning, resulting in consistently improved torque prediction accuracy across all evaluated models and demonstrating its suitability for modeling flexible robotic systems. Among the evaluated approaches, the gray-box physics-based regression model achieved the best overall performance, while the black-box and purely physics-based models exhibited larger baseline errors. Nevertheless, the inclusion of Gaussian mixture regression substantially improved torque prediction accuracy for all models, highlighting the effectiveness of residual learning in compensating for unmodeled flexible dynamics.

\bibliographystyle{ieeetr}
\bibliography{samplebib}

\end{document}